\documentclass[a4paper,fleqn]{cas-dc}

\usepackage[authoryear,longnamesfirst]{natbib}

\usepackage{pgfplots}
\pgfplotsset{compat=1.17}

\def\tsc#1{\csdef{#1}{\textsc{\lowercase{#1}}\xspace}}
\tsc{WGM}
\tsc{QE}
\tsc{EP}
\tsc{PMS}
\tsc{BEC}
\tsc{DE}
\begin{document}
\let\WriteBookmarks\relax
\def\floatpagepagefraction{1}
\def\textpagefraction{.001}

\shorttitle{Which Student is Best?}
%\shorttitle{SMOLTURTL}

\shortauthors{Nityasya, et al.}

%\title [mode = title]{SMOLTURTL \emoji{turtle}: Small Model Tutelage via Task-Specific BERT Distillation}
\title [mode = title]{Which Student is Best? A Comprehensive Knowledge Distillation Exam for Task-Specific BERT Models}

%\tnotemark[1,2]

%\tnotetext[1]{This document is the results of the research
%   project funded by the National Science Foundation.}

%\tnotetext[2]{The second title footnote which is a longer text matter
%   to fill through the whole text width and overflow into
%   another line in the footnotes area of the first page.}

% \author[1]{CV Radhakrishnan}[type=editor,
%                         auid=000,bioid=1,
%                         prefix=Sir,
%                         role=Researcher,
%                         orcid=0000-0001-7511-2910]


% % Footnote of the first author

% % Email id of the first author

% % URL of the first author
% \ead[url]{www.cvr.cc, cvr@sayahna.org}

% %  Credit authorship
% \credit{Conceptualization of this study, Methodology, Software}

\author[1]{Made Nindyatama Nityasya}[orcid=0000-0002-5570-4676]
\ead{made@kata.ai}
% \cormark[1]

% Other authors
\author[1]{Haryo Akbarianto Wibowo}[orcid=0000-0001-7457-2165]
\ead{haryo@kata.ai}

\author[1]{Rendi Chevi}[orcid=0000-0002-7917-6556]
\ead{rendi.chevi@kata.ai}

\author[1,2]{Radityo Eko Prasojo}[orcid=0000-0002-5148-7299]
\ead{ridho@kata.ai}

\author[1]{Alham Fikri Aji}[orcid=0000-0001-7665-0673]
\ead{aji@kata.ai}

\affiliation[1]{organization={Kata.ai Research Team},
    city={Jakarta},
    country={Indonesia}}
    
\affiliation[2]{organization={University of Indonesia},
    city={Depok},
    country={Indonesia}}

% \nonumnote{This note has no numbers. In this work we demonstrate $a_b$
%   the formation Y\_1 of a new type of polariton on the interface
%   between a cuprous oxide slab and a polystyrene micro-sphere placed

\begin{abstract}
We perform knowledge distillation (KD) benchmark from task-specific BERT-base teacher models to various student models: BiLSTM, CNN, BERT-Tiny, BERT-Mini, and BERT-Small. Our experiment involves 12 datasets grouped in two tasks: text classification and sequence labeling in the Indonesian language. We also compare various aspects of distillations including the usage of word embeddings and unlabeled data augmentation. Our experiments show that, despite the rising popularity of Transformer-based models, using BiLSTM and CNN student models provide the best trade-off between performance and computational resource (CPU, RAM, and storage) compared to pruned BERT models. We further propose some quick wins on performing KD to produce small NLP models via efficient KD training mechanisms involving simple choices of loss functions, word embeddings, and unlabeled data preparation.
\end{abstract}

% Keywords

\maketitle

\section{Introduction}

In recent years, Deep Learning (DL) models have been widely utilized to solve Natural Language Processing (NLP) problems like text classification, sequence labeling, and machine translation. Given a large amount of labeled data, DL can train a highly-generalizable model thanks to its internal complexity and depth.

In NLP, the data (or lack thereof) itself is an infamous problem, especially in low-resource languages. The current common solution to this problem is by (1) \textbf{pretraining} -- to perform a large-scale representation learning of textual data, leveraging the abundance of unlabeled text data from the Internet, and then (2) \textbf{fine-tuning} -- to carry the parameters learned from the representation learning step and continuing training with the real labeled data on the specific problem we want to solve. The general, intermediate model as a result of the pretraining step is usually called a language model (LM). Pretraining an LM typically requires significantly more data and computational resources than fine-tuning it, and therefore takes longer. However, an LM can be fine-tuned into several models serving different purposes, therefore potentially reducing the overall training time and cost.

Due to this benefit, several studies have focused on improving the pretraining step. To this end, there is a trend of rising size and complexity of the resulted language models, starting from distributed representation learning of words like word2vec~\citep{mikolov2013distributed} and GloVe~\citep{pennington2014glove}, subword information enrichment~\citep{bojanowski2017enriching}, contextual representation using Recurrent Neural Networks (RNNs)~\citep{Peters:2018}, and finally the Transformer-based models~\citep{vaswani2017attention} that ranges from hundreds of millions to billions of trainable parameters.

Transformer~\citep{vaswani2017attention}, like RNN, uses an encoder-decoder architecture and therefore is originally designed for sequence-to-sequence learning like machine translation. However, previous work discovered several ways to utilize Transformer to train language models, such as (1) autoregressive models trained using only the decoder blocks of Transformer, like GPT~\citep{radford2018improving}, GPT-2~\citep{radford2019language}, and GPT-3~\citep{brown2020language}, (2) masked-language models using only the encoder blocks, like BERT~\citep{devlin2019bert} and its variants such as RoBERTa~\citep{liu2019roberta} and ELECTRA~\citep{duan2020electra}, and (3) a combination of both, like BART~\citep{lewis2020bart} and T5~\citep{raffel2019exploring}. There is a similar trend that further fosters the notion of ``bigger is better'' in Transformer-based LMs. Evidently, common benchmark such as GLUE~\citep{wang2018glue} shows that the larger version of a model (e.g. BERT-large; GPT-3) outperforms the smaller ones (e.g. BERT-base and BERT-small; GPT-2 and GPT) almost universally in terms of several performance metrics such as Precision, Recall, and F1-score on various NLP tasks.

Naturally, training and running larger models bear additional costs in terms of computational power. The costs mainly concern the Graphical Processing Unit (GPU), Tensor Processing Unit (TPU), or Central Processing Unit (CPU) used for training and inference, as well as the model size in terms of parameter size or memory requirement.
Unfortunately, these costs are often not discussed in various model benchmarks, leaving the cost-performance trade-off unclear~\citep{nityasya2021costs}.

Typically, modern DL models require GPUs or TPUs to train, while using CPUs is generally considered infeasible, especially for large models with many fully connected (FC) layers~\citep{wang2019benchmarking} like Transformer. On the other hand, running the model for inference on CPU is feasible to some extent, that is, until scalability becomes a problem. An NLP model can be deployed within applications such as chatbots or search engines serving a large user base. In this case, using small classical machine learning (ML) models makes more sense than using big Transformer-based models~\citep{nityasya2021costs} due to the scalability issue, while sacrificing accuracy. Still, depending on the difficulty of the NLP task, a classical ML model might not be powerful enough to produce consistently good results compared to a Transformer-based model. Besides scalability, another issue might arise when we want to deploy a large Transformer-based model on low-powered devices such as smartphones~\citep{sun2020mobilebert} or other embedded devices, where storage and CPU resource is at a minimum.

In this work, our general theme is finding a middle ground: can we build a small model close to the size of a typical classical ML model but with a great performance close to a typical Transformer-based model? More specifically, we focus on \textbf{Knowledge Distillation} (KD) approach~\citep{ganesh2021compressing,kim2016sequence,tang2019distilling,sanh2019distilbert,raffel2019exploring,jiao-etal-2020-tinybert,sun2020mobilebert,chen2020adabert,he2021distiller} leveraging student-teacher models, that is to train smaller models (student) from a large model (teacher), resulting in students with similar performance to the teacher. Our contribution is as follows:
\begin{itemize}
    \item We perform KD from BERT-based teacher models fine-tuned on various tasks into several student models based on RNNs, Convolutional Neural Networks (CNNs), and Transformers. In the spirit of efficiency and reproducibility, we distill fine-tuned models instead of pretrained models because the former is much less resource-intensive.
    \item We perform ablation studies on each student model by employing embedding weight transfer, leveraging data augmentation, and converting the final model into an optimized ONNX\footnote{https://github.com/onnx/onnx} version for the final inference test.
    \item We compare the performance of the student models in terms of performance, running time, and memory/storage footprint. To the best of our knowledge, our work is the first to provide a comprehensive practical distillation benchmark of various student models in various tasks.
    \item Our benchmark is done for various text classification and sequence labeling tasks in the Indonesian language. To the best of our knowledge, we are the first to develop and benchmark various distilled models in Indonesian. We argue that our approach can be generalized into any other language, provided that a teacher model and some labeled data for benchmarking exist in that language. We leave the benchmark for the sequence-to-sequence tasks as future work.
\end{itemize}

Based on our experiment, we found that the best distilled BiLSTM model has the size of only 3\% of the original BERT and run 22x times faster in CPU machine with only 3-4 point of performance difference in F1-score. A better performed distilled BERT-Mini model has the size of 9\% the BERT model and runs 10x faster. Converting the BiLSTM model to an ONNX further boosts the inference time into 100x faster than the BERT model, while having 2.5\% of the size.
%Furthermore, we are able to compress the model even more by using a technology called ONNX and achieve the final model of 60\% of the original non ONNX model and able to speed up the performance up to 4x time on CPU.
At the end, we also perform several additional experiments to discuss some aspects affecting the distillation performance, and then propose some suggestions to improve the overall KD results involving domain, data balancing, and the length of unlabeled data augmentation, as well as hyperparameter tuning.

\section{Related Work}
\label{sec:related-work}

The computational cost of running deeper neural models motivates the quest towards building smaller models at a low cost. One relevant topic, model compression, focuses on making smaller and faster models for inference while keeping performance loss at minimum. There are several ways to do the model compression, such as pruning, quantization~\citep{wu2018training}, and knowledge distillation (KD).

KD was first introduced by~\citet{hinton2015distilling} to combat the need for multiple ensemble models and was shown to perform with great results on MNIST dataset. In NLP, KD had been used to compress large models for machine translation~\citep{kim2016sequence}, classification~\citep{tang2019distilling}, and sequence labeling~\citep{tsai2019small}. Following the trend of pretrained LMs, KD has also been performed to make smaller general-purpose language models like DistilBERT~\citep{sanh2019distilbert} and MobileBERT~\citep{sun2020mobilebert}. These pre-trained models can be fine-tuned for a specific task, just like how a regular pre-trained model can.

KD can also be performed on fine-tuned models, which is widely known as task-specific distilation~\citep{tang2019distilling,adhikari-etal-2020-exploring,turc2019,hagstrom-johansson-2021-knowledge}. We first train a teacher model by fine-tuning a pre-existing language model over a task-specific dataset, then train the students from the teacher's output. Distilling fine-tuned models this way is less computationally demanding compared to distilling general-purpose language models. Despite both having a smaller model size, the latter still requires lots of training data due to the nature of pretraining. In contrast, distilling fine-tuned models require much smaller training data. Therefore, in this work, we mainly focus on task-specific distillation.

\begin{table*}[]
    \centering
    \caption{Comparison between our work and previous task-specific KD studies in NLP}
    \resizebox{\textwidth}{!}{\begin{tabular}{|l|c|c|c|c|}
        \toprule
         Work & Student models & Tasks & Language & \#Datasets \\ \midrule
         \citet{adhikari-etal-2020-exploring} & CNNs, LSTMs & Classification & English & 4 \\
         \citet{tang2019distilling} & LSTMs & Classification & English & 4 \\
         \citet{turc2019} & Pruned Transformers & Classification & English & 6 \\
         \citet{hagstrom-johansson-2021-knowledge} & CNNs, LSTMs & Sequence Labeling & Swedish & 1 \\
         \midrule
         Ours & CNNs, LSTMs, pruned Transformers & Classification, Sequence Labeling & Indonesian & 12 \\ 
         \bottomrule
    \end{tabular}}
    \label{tab:comparison}
\end{table*}

The work by \citet{adhikari-etal-2020-exploring} (See Table~\ref{tab:comparison}) is the most similar to ours, although we extend the distillation benchmark to include also small Transformers on top of LSTMs and CNNs for the student models. Moreover, we also perform experiments on some sequence labeling tasks on top of the classification tasks. \citet{tang2019distilling} and \citet{hagstrom-johansson-2021-knowledge} performed similar distillation benchmark, but focused on a smaller scope. \citet{turc2019} focused on pruned Transformer models, specifically on smaller-sized BERTs. They recommend the student models to be pretrained before the distillation process, as this boosts the final results. However, because it is computationally demanding, we forgo student pretraining in this work.

Recently, \citet{jiao-etal-2020-tinybert} developed TinyBERT, which uses a two-staged learning framework where they performed distillation on both the pre-training and fine-tuning stages. As the two stages are orthogonal, it is possible to leverage its fine-tuning steps for task-specific distillation. However, we leave exploration in this direction for future work.

\section{Methodology}
\subsection{Knowledge Distillation}

Knowledge Distillation (KD) is a technique to transfer the knowledge from a cumbersome models to a new simpler model~\cite{hinton2015distilling}.
The cumbersome model, also known as teacher model, usually has a lost of parameter or is an ensemble of models which is difficult to deploy on device with a limited computing power. Meanwhile, the new simpler model, called the student model, usually has fewer parameters, hence more suitable for deployment.
The goal of KD is to extract information from the teacher model to the student model
while removing the unneeded information. 
The student try to mimic the teacher's behaviour by producing similar outputs, thus gaining a performance similar to the teacher. 
The student model is trained to minimize the KD loss, that is, the difference between the teacher's output and student's output.

%There are two well-known approaches to determine the KD loss. The approaches are either using the hard target or the soft target to compute the KD loss.
There are two common ways to compute KD loss: hard target and soft target. For hard target, we use the output labels from the teacher as the ground truth for the student model to compare its own output labels with. 
% This approach is easier to implement especially when we use a non-neural model as the student.
In soft target, we compare the error between the output logits or the probability distribution from the softmax output between the student and the teacher. Table~\ref{tbl-hard-soft-example} shows some examples between hard target and soft target.

\begin{table}[pos=h]
\caption{Difference between using hard target and soft target, using sentiment analysis task as example. Hard target are labels, while soft target are their probability distribution given by the teacher model.}
\label{tbl-hard-soft-example}
\begin{tabular}{|l|c|c|}
\toprule
\multicolumn{1}{|c|}{Input} & hard target & soft target   \\ 
\midrule
I can see the sunset   & neutral  & {[}0.3 \textbf{0.5} 0.2{]} \\
The room feels fresh    & positive & {[}\textbf{0.7} 0.2 0.1{]} \\
The food is very spicy & negative & {[}0.3 0.3 \textbf{0.4}{]} \\
\bottomrule
\end{tabular}
\end{table}

While previous KD studies have compared the use of soft target, hard target, and their combinations~\cite{yang2021knowledge}, more recent work~\citep{KimOKCY21comparing} has also compared two different soft target losses: Mean Squared Error (MSE) and Kullback-Leibler Divergence (KLD). MSE is used to calculate the loss between the logits. It averages the square difference (error) between the teacher logits and student logits (Equation~\ref{eq_mse}). Meanwhile, KLD is used to calculate the loss between two probability distributions. Specifically, KLD measures relative entropy between the two distributions as the average difference of their log probabilities (Equation~\ref{eq_kld}). \citet{KimOKCY21comparing} results suggest that MSE is superior, though prior to that, KLD is more generally used.

\begin{equation} \label{eq_mse}
\text{MSE} = \displaystyle\frac{1}{n}\sum_{t=1}^{n}(x_i-y_i)^2
\end{equation}
\begin{equation} \label{eq_kld}
\text{KLD} = \displaystyle\frac{1}{n}\sum_{t=1}^{n}y_i.(\log(y_i-x_i))
\end{equation}

In the original implementation oF KD leveraging soft target~\citep{hinton2015distilling}, a hyperparameter called temperature $T$ is used to divide the output logits $z_i$ before passing it to the softmax function. This is done to make a softer probability distribution output $q_i$ while retaining the relativeness\footnote{that is, the sum of the distribution remains equal to 1} of the output~\citep{hinton2015distilling}. 
\begin{equation} \label{eq_soft}
q_i = \displaystyle\frac{\exp(z_i/T)}{\sum_{j}\exp(z_j/T)}
\end{equation}
%In a recent NLP work, they found that temperature yields the best result if we set the temperature hyperparameter to 1~\citep{jiao-etal-2020-tinybert}.
In a separate study, \citet{jiao-etal-2020-tinybert} found that KD in NLP tasks yield the best results when the temperature value is set to 1 (tested on GLUE~\cite{wang2018glue} datasets), which we follow throughout our experiment.

%Some distilled architecture calculate student's output with the original label, usually called student loss. The final loss is the combination between student loss and the distil loss. One drawback for this architecture is that we need a labelled data when we want to try KD with additional data.

Beside soft target vs hard target, recent studies have explored various other losses, e.g. incorporating student loss using labeled data and 
comparing logits from layers before the output layer~\citep{yang2021knowledge}. We leave exploration in this direction as future work.

%Furthermore when we use unlabelled data as training a KD model, we do need to annotated those data as the original label would not be used.

\subsection{Model}

We use fine-tuned BERT-base models as our teacher model. For the student, we used several smaller neural models such as Bi-LSTM, CNN, and some pruned BERT architecture. 

\subsubsection{BERT}

BERT~\citep{devlin2019bert} is a Transformer-based model~\citep{vaswani2017attention} that has since spawned several variations that often would top the chart of NLP benchmarks such as GLUE~\citep{wang2018glue}. BERT owes its great performance from its size and complexity, where it is able to accurately model a large amount of text data within its millions of parameters. As such, a BERT-based model is a suitable candidate for a teacher model due to its performance and sheer size.

In this work, we use fine-tuned BERT-based models as the teacher models. To obtain these teacher models, we fine-tune IndoBERT,\footnote{\url{https://huggingface.co/indobenchmark/indobert-base-p1}} an Indonesian pretrained language model from IndoNLU~\citep{wilie2020indonlu}, to several task-specific models using some labelled data. IndoBERT itself was trained on a large corpus of Indonesian language, containing around 240M sentences, and was shown to have a great performance when tested on IndoNLU's benchmark~\citep{wilie2020indonlu}. 

\subsubsection{LSTM}

Long Short Term Memory (LSTM) is a variant of Recurrent Neural Network (RNN), which is widely used for sequence modeling such as in time-series analysis and NLP. Compared to vanilla RNN, LSTM has an additional feature called gating mechanism to filter the information to be passed to the next step, yielding better performance. Prior to Transformer, LSTM-based RNN models are often considered as the state-of-the-art neural models for solving NLP problems~\citep{huang2015bidirectional,chen2017improving}. Coupled with its relatively-smaller size compared to Transformer-based models such as BERT, we consider LSTM as a great option for a student model.
%Many NLP researchers use LSTM to benchmark and evaluate their system. Thus, we also compare LSTM with other system. % katanya pperlu dibenerin

\begin{figure}
  \centering
  \caption{Our LSTM implementation with self-attention. The implementation of the attention layer is based on  \cite{vaswani2017attention}.}
  \includegraphics[width=.9\linewidth]{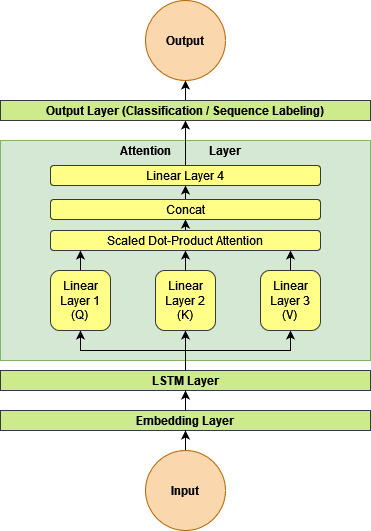}
  \label{fig:lstm}
\end{figure}

Our LSTM student model consists of Bi-directional LSTM (Bi-LSTM), where the architecture will consider the left-to-right direction sequence of input and the opposite one. 
%To make the model have more capability on memorizing more longer sentence, 
We add an attention mechanism to the model so that the model does not consider all tokens to be equal in value, i.e. it pays `more attention' to the more important tokens by learning different weights over the tokens in the sequence. 
% Previous work has added attention mechanisms into Bi-LSTM in various different ways~\citep{bahdanau2014neural,luong2015effective} and has been shown to significantly increase performance. 
In this work, we use the self-attention mechanism from Transformer itself~\citep{vaswani2017attention}, which is one of the  attention mechanism often used in sequential deep learning model. We plug it on top of the Bi-LSTM layers. Refer to Figure \ref{fig:lstm} to see more details about the implementation.

To get the model's logits, we put a linear layer to map the output of the LSTM model to its respective's output. For the classification task, we follow BERT's~\citep{devlin2019bert} implementation to take the vector of the first sequence of the output, that represents  the start of token, as the input of the linear layer. In the Token Labeling task, we use all of the sequence vector as the input of the linear layer.

\subsubsection{CNN}

CNN is widely used in image processing. However, by changing the 2D convolution into 1D, it could be applied also for text data. CNN has been shown to have good results in text classification~\citep{kim-2014-convolutional}. 

Our CNN student model is composed of a backbone with stacks of residual convolutional network and classifier head for each task. The backbone starts with a word embedding layer, we apply absolute positional encoding~\citep{vaswani2017attention} to inject positional information to the input texts. Then, we feed the embedding vectors into N-stacks convolution layers with residual connection~\citep{he2016deep}. We use depth-wise separable convolution~\citep{chollet2017xception} to keep the student's size relatively small and layer normalization 
%[\href{https://arxiv.org/abs/1607.06450}{Link}] 
as the activation function. 

For the sequence labeling task, we directly feed the backbone's output into a linear layer. 
For the classification task, we apply global average pooling~\citep{lin2013network} to the backbone's output followed by a linear layer to obtain the output logits. 

\subsubsection{Pruned Transformers: BERT-Tiny, Bert-Mini, and BERT-Small}

As we still want to leverage the powerful architecture of a Transformer-based model, we tried to use some of the proposed smaller size BERTs by~\citet{turc2019}: BERT-Tiny, BERT-Mini, and BERT-Small, each having different sizes of attention heads, transformer layers, and embedding sizes, hence different total numbers of parameters as well (Table~\ref{tbl-bert-compare}). We note again, as we did in Section~\ref{sec:related-work}, that we leverage these small BERT models for KD without first pretraining them to focus on more efficient KD approaches.

\begin{table}[pos=h]
\caption{Differences between BERT-base, BERT-small, BERT-Mini, and BERT-Tiny in terms of their attention heads, transformer layers, embedding sizes, and numbers of parameters.}
\label{tbl-bert-compare}
\begin{tabular}{|l|cccc|}
\toprule
BERT model & \#Attn & \#TLayer & \#Embed & \#Param \\
\midrule
BERT-base & 12 & 12 & 768 & 110.1M \\
BERT-small & 8 & 4 & 512 & 29.1M \\
BERT-mini & 4 & 4 & 256 & 11.3M \\
BERT-tiny & 2 & 2 & 128 & 4.4M \\
\bottomrule
\end{tabular}
\end{table}

%The configuration for BERT-Small\footnote{https://github.com/google-research/bert} contains 8 attention head (A), 4 transformer layer (L), and 512 embedding size (H). The total number of parameter is only 29.1M, a lot smaller model in compare to BERT-Base which has 110.1M. 

% We also conduct an exploration with BERT-Tiny's configs, but the final result is no better than CNN or Bi-LSTM model even after KD. 

\subsection{Unlabelled data augmentation}
\label{sec-ulb}

Adding more data during KD training can improve the overall performance~\citep{papernot2016semi,vongkulbhisal2019unifying}.
%Using either a labelled data or unlabelled data does not differ too much as we only interest in the teacher's output with no regard to the actual labels.
The data does not need to be labelled because we leverage the teacher's model output instead. In this work, we will explore the effect of unlabeled data augmentation for each of the student models.

%In this experiment, we simply combined the unique training data from all dataset. By doing this, we also want to now the effect of adding out-of-domain data to KD's performance.

\subsection{Word Embeddings}

Pretrained word embeddings enable feature extraction to initialize the weights of the embedding layer of an NLP model. This has been empirically proven to boost model performance in several NLP tasks~\citep{sachan2019lstm,qi-etal-2018-pre,howard-ruder-2018-universal,dai2015seq} and to help the model converges faster. Using word embeddings also serves as a cheaper alternative to student pretraining prior to KD, a recommended step~\citep{turc2019} we skip in this work due to computational cost consideration.

%We also found that training the student model often took a longtime into convergence~(See Appendix Figure~\ref{fig:train-converge-embed}). 

%To speed up training convergence and boost student's performance, we experimented to initialize the student's word embeddings with pre-trained embeddings instead of initialize it randomly.

In general, there are two types of word embeddings: non-contextual and contextual. In contextual embedding, a word can have several embeddings depending on the context (that is, the other words around it), whereas it can only have one in non-contextual embedding. Among the earliest non-contextual embeddings are word2vec~\citep{mikolov2013distributed} and GloVe~\citep{pennington2014glove}, while fasttext~\citep{bojanowski2017enriching} is a later improvement which utilizes subword embeddings to handle out-of-vocabulary. Meanwhile, contextual embeddings are typically obtained from the embedding layer\footnote{though to fully obtain the contextuality BERT, embeddings can also be obtained from the middle or output layers~\citep{rogers2020primer}; we leave exploration in this direction as future work.} of RNN-based or Transformer-based pretrained language models, such as ELMO~\citep{Peters:2018} and BERT~\citep{devlin2019bert}. In this work, we explore the effect of leveraging contextual vs non-contextual word embeddings for KD.

% We tried two different pre-trained word embeddings: FastText and BERT embedding. 
% For the sFastText embeddings, we use the 300-dimensional word embeddings pre-trained on the Indonesian Common Crawl dataset~\citep{grave2018learning}. Whereas for BERT embedding, we use the embedding layer from the IndoNLU BERT pre-trained model ~\citep{wilie2020indonlu} which has the dimension of 768.
% Because we set the default embedding size of CNN and BiLSTM students to 100-dimension only, we reduce the dimensionality of the pre-trained embeddings weights with SVD (Singular Value Decomposition) to initialize both students.

\section{Experiment}
In this section, we first discuss the datasets that we use for experiment. Then, we describe the general experiment setup, the models that we train for the experiments, and the metrics that we use. 

\subsection{Dataset}

We use six datasets for each task. The description of each dataset can be seen below in Sections~\ref{sec:classification-data} and~\ref{sec:seqlab-data}, while the statistics can be seen in Table~\ref{tbl-data}.
For EmoT, NERGrit, NERP, and POSP dataset, we use the same split as \citeauthor{wilie2020indonlu}'s (\citeyear{wilie2020indonlu}), while we perform our own split for the other datasets.

\begin{table}[width=.9\linewidth,cols=5,pos=h]
\caption{Data statistic. Number of train, dev, and test data. For sequence labeling, number of classes was count without BIO.}
\label{tbl-data}
\begin{tabular*}{\tblwidth}{LRRRR}
\toprule
Dataset    & Train  & Dev   & Test & \#class \\ \midrule
\multicolumn{5}{c}{Text Classification Task} \\ \midrule
Smltk      & 1134   & 1280  & 1272 & 96      \\
Health     & 57938  & 6894  & 6897 & 5       \\
Telco      & 11520  & 1440  & 1440 & 144     \\
Sentiment  & 3638   & 399   & 1011 & 2       \\
SMSA       & 11000  & 1260  & 500  & 3       \\
EmoT       & 3521   & 440   & 442  & 5       \\ \midrule
\multicolumn{5}{c}{Sequence Labeling Task}   \\ \midrule
EntK       & 10955  & 1250  & 1372 & 14      \\
TermA      & 3000   & 1000  & 1000 & 3       \\
POS        & 7222   & 802   & 2006 & 23      \\
NERGrit    & 1672   & 209   & 209  & 4       \\
NERP       & 6720   & 840   & 840  & 5       \\
POSP       & 6720   & 840   & 840  & 26      \\ 
\bottomrule
\end{tabular*}
\end{table}

\subsubsection{Text Classification Tasks}
\label{sec:classification-data}

\medskip

\noindent\textbf{Smltk} -- A dataset containing intent classification for small talk conversations with chatbots in Indonesian. The language used is informal and the data is imbalanced. Example of classes: askJokes, askTime, greetings, askWeather, etc.

\medskip

\noindent\textbf{Health} -- A text classification dataset for an online medical consultant. It contains conversation between doctor and patient which grouped into complaint, patient's actions, diagnosis, recommendation, and fallback. 

\medskip

\noindent\textbf{Telco} -- A semi-formal classification dataset for questions and tasks that can be answered automatically by a telecommunication’s bot. Example of the classes: ask promo, error network, postpaid registration, 4G activation, etc.

\medskip

\noindent\textbf{SentA} -- A binary classification sentiment analysis dataset that contains text from Twitter and hotel review, obtained from IndoLEM~\citep{koto2020indolem}. 

\medskip

\noindent\textbf{SMSA}\footnote{\url{https://huggingface.co/datasets/indonlu}} -- A sentiment analysis dataset that that is grouped into three classes: positive, negative, and neutral. It contains comments and reviews from various Indonesian online platforms~\citep{purwarianti2019improving}.

\medskip

\noindent\textbf{EmoT}\footnotemark[5] -- A Twitter emotion classification dataset that was grouped into 5 classes: anger, fear, happiness, love, and sadness~\citep{saputri2018emotion}.

\subsubsection{Sequence Labeling Tasks}
\label{sec:seqlab-data}

\medskip

\noindent\textbf{EntK} -- A manually gathered and annotated NER dataset that contains fourteen different labels: person, location, email, number, phone,ur date-time, currency, and 5 different units.

\medskip

\noindent\textbf{TermA}\footnote{\url{https://github.com/jordhy97/final\_project}} -- An aspect and opinion term extraction dataset from a hotel aggregator platform, AiryRooms, that contains sentiment for each aspect~\citep{fernando2019aspect}.

\medskip 

\noindent\textbf{POS}\footnote{\url{https://github.com/kmkurn/id-pos-tagging}} -- POS Tagging dataset that was collected from PAN Localization Project~\citep{dinakaramani2014designing}. We follow the data split provided by by~\citet{kurniawan2018toward}.
\medskip

\noindent\textbf{NERGrit}\footnotemark[5] -- A NER dataset from Grit-ID repository.\footnote{\url{https://github.com/grit-id/nergrit-corpus}} The tags are person, location, organization in IOB format.\footnote{A common tagging format that indicate a chunk of text, using prefix B- as Begin and I- as Inside~\citep{ramshaw-marcus-1995-text}}

\medskip

\noindent\textbf{NERP}\footnotemark[5] A NER dataset containing articles from several Indonesia news website~\citep{hoesen2018investigating}. The tags are PER (person), LOC (location), IND (product or brand), EVT (event), FNB (food and beverages) in its IOB format.

\medskip

\noindent\textbf{POSP}\footnotemark[5] A POS Tagging dataset with the same data as NERP. It contains 26 POS tag classes that follows
Indonesian Association of Computational Linguistics
(INACL).\footnote{\url{http://inacl.id/inacl/wp-content/uploads/2017/06/INACL-POS-Tagging-Convention-26-Mei.pdf}}

\subsection{Models to Train}

In each task, we fine-tune a BERT-base model as the general baseline. Then, in a hill-climbing setting, we train several other models:
\begin{enumerate}
    \item \textbf{Vanilla} -- vanilla student models, that is, ``student'' models trained without involving KD, using only the labeled data.
    \item \textbf{KD} -- student models trained with KD using the BERT-base model as a techer. We use MSE loss over KLD following~\citet{KimOKCY21comparing}.
    \item \textbf{KD Ulb} -- KD models that are improved by using unlabeled data.
    \item \textbf{KD Ulb + Word Embeddings} -- KD models with unlabeled data that are trained leveraging pre-trained word embeddings. We compare two embeddings: a non-contextual one, FastText, and a contextual one, BERT.  
\end{enumerate}

Our objective is to observe whether the use of KD, the use of unlabeled data during KD, and the use of pretrained embedding (contextual vs non-contextual) affect the performance of our models. Later on, we transform the best performing models into an optimized ONNX version for inference tests on CPU. We compare the above models using several metrics to find the trade-offs between their performance and computational cost.

\subsection{Metrics}

We use macro F1 Score to measure the performance for both text classification and sequence labeling task. To calculate the cost of inference, we kept track of the model size and inference time. For the model size, we compare the number of each model's parameters. For inference time, we compute the average time needed to run a single sentence inference over multiple iterations with various sentence length. %These metrics can be used to provide a broader trade-off between performance and cost.
 
\subsection{Experiment Setup}

All experiments was trained on a single GPU RTX 3090. %The batch size we use for finetuning is either 16 or 32~\citep{devlin2019bert} while for distilation it is 32 or 64.
%We use AdamW~\citep{loshchilov2017decoupled} optimizer for text classification, while for sequence labeling we use Adam~\cite{kingma2014adam} optimizer.
We set a very high number for number of training epoch but applied an early stopping mechanism with the patience of 10 based on validation loss. For learning rate, we choose the best from [5e-3, 1e-3, 5e-4, 1e-4, 5e-5, 1e-5]. For CPU inference tests, we use a machine with a 6-core Intel i5 CPU and 24 GB of RAM.

\section{Results}
Across the board, the BERT-base teacher model outperforms the other models, which we discuss thoroughly below. We first focus on the models' performance in terms of F1 score (see Table~\ref{tbl-cls} and Table~\ref{tbl-seqlab} for the complete results of classification and sequence labeling tasks, respectively). Then, we discuss the models' cost analysis in terms of their size and inference time (Table~\ref{tbl-cost} and Table~\ref{tbl-onnx}).

\subsection{Model's Performance}

In this section, we discuss the model's performance on each experiment setting. We start from the baseline experiment then continue with the KD experiment and several additional experiments to further improve the KD's performance.

\begin{table*}[]
\caption{Classification KD}
\label{tbl-cls}
\begin{tabular}{lrrrrrrr}
\toprule
\multicolumn{1}{c}{Models} & \multicolumn{1}{c}{Smltk} & \multicolumn{1}{c}{Health} & \multicolumn{1}{c}{Telco} & \multicolumn{1}{c}{SentA} & \multicolumn{1}{c}{EmoT} & \multicolumn{1}{c}{SMSA} & \multicolumn{1}{c}{avg} \\ \midrule
BERT              & 96.83 & 80.01 & 94.96 & 90.01 & 72.30 & 91.49 & 87.60 \\ \midrule
Bi-LSTM           & 86.78 & 72.22 & 88.16 & 76.35 & 55.96 & 83.84 & 77.22 \\
KD                & 92.96 & 74.25 & 91.41 & 82.28 & 66.72 & 88.67 & 82.72 \\
KD Ulb            & 92.74 & 77.61 & 92.71 & 85.17 & 68.33 & 88.81 & 84.23 \\ 
KD Ulb + FastText & 93.03 & 77.97 & 92.57 & 85.80 & 69.40 & 88.34 & 84.52 \\
KD Ulb + BERT Embed & 92.79 & 78.78 & 92.15 & 86.28 & 69.54 & 88.77 & 84.72 \\
\midrule
CNN               & 88.24 & 71.97 & 88.22 & 77.95 & 57.57 & 84.12 & 78.01 \\
KD            & 92.43 & 69.85 & 90.95 & 80.33 & 63.90 & 87.26 & 80.79 \\
KD Ulb        & 84.50 & 74.32 & 84.05 & 82.46 & 63.92 & 87.75 & 79.50 \\  
KD Ulb + FastText & 81.23 & 73.27 & 86.40 & 84.65 & 65.97 & 88.44 & 79.99 \\
KD Ulb + BERT Embed & 85.35 & 74.20 & 86.72 & 84.01 & 64.48 & 87.40 & 80.36 \\
\midrule 
BERT-Tiny & 89.89 & 69.46 & 86.31 & 77.36 & 60.42 & 79.88 & 77.22 \\
KD & 91.87 & 76.27 & 90.05 & 81.56 & 67.10 & 83.79 & 81.77 \\
KD Ulb & 94.02 & 77.74 & 92.17 & 83.06 & 66.21 & 86.39 & 83.26 \\
KD Ulb + FastText & 94.48 & 78.32 & 92.40 & 82.52 & 66.97 & 86.90 & 83.60 \\
KD Ulb + BERT Embed & 95.31 & 77.02 & 92.69 & 83.17 & 66.90 & 86.28 & 83.56 \\
\midrule
BERT-Mini & 91.62 & 68.88 & 89.70 & 80.47 & 60.27 & 81.93 & 78.81 \\
KD & 94.40 & 77.95 & 92.96 & 81.78 & 68.26 & 86.81 & 83.69 \\
KD Ulb & 95.10 & 77.87 & 93.43 & 83.57 & 69.62 & 85.02 & 84.10 \\
KD Ulb + FastText & 96.37 & 77.93 & 93.72 & 83.95 & 68.87 & 86.10 & 84.49 \\
KD Ulb + BERT Embed & 95.73 & 77.95 & 94.22 & 81.95 & 69.75 & 87.99 & 84.60 \\
\midrule
BERT-Small        & 92.92 & 73.81 & 91.35 & 80.20 & 62.65 & 83.77 & 80.78 \\
KD     & 94.74 & 78.82 & 92.69 & 80.78 & 69.01 & 87.21 & 83.88 \\
KD Ulb & 96.51 & 78.73 & 94.18 & 83.90 & 69.60 & 87.51 & 85.07 \\  
KD Ulb + FastText & 96.38 & 77.94 & 94.25 & 82.43 & 69.83 & 87.85 & 84.78 \\
KD Ulb + BERT Embed & 96.30 & 77.95 & 94.54 & 85.16 & 70.53 & 88.32 & 85.47 \\
\bottomrule
\end{tabular}
\end{table*}

\begin{table*}[]
\caption{Sequence Labelling KD}
\label{tbl-seqlab}
\begin{tabular}{lrrrrrrr}
\toprule
\multicolumn{1}{c}{Models} & \multicolumn{1}{c}{EntK} & \multicolumn{1}{c}{TermA} & \multicolumn{1}{c}{POS} & \multicolumn{1}{c}{NERGrit} & \multicolumn{1}{c}{NERP} & \multicolumn{1}{c}{POSP} & \multicolumn{1}{c}{avg} \\ \midrule
BERT                & 90.86 & 90.38 & 95.66 & 75.78 & 78.35 & 95.81 & 87.81 \\ \midrule
BiLSTM              & 79.21 & 84.38 & 91.12 & 44.00 & 58.56 & 93.59 & 75.14 \\
% KD MSE Loss         & 83.32 & 87.13 & 93.05 & 52.56 & 66.88 & 94.95 & 79.65 \\
KD                  & 85.99 & 88.08 & 93.56 & 56.80 & 69.10 & 95.05 & 81.43 \\
KD Ulb              & 86.24 & 87.82 & 94.11 & 62.90 & 70.38 & 95.57 & 82.84 \\
KD Ulb + FastText   & 86.50 & 88.15 & 94.29 & 66.52 & 72.27 & 95.70 & 83.91 \\
KD Ulb + BERT Embed & 86.28 & 88.22 & 94.44 & 68.26 & 72.40 & 95.77 & 84.23  \\ \midrule
CNN                 & 80.17 & 84.70 & 91.94 & 39.86 & 59.52 & 94.45 & 75.11 \\
KD                  & 87.41 & 88.63 & 93.97 & 60.10 & 71.14 & 94.95 & 82.70 \\
% KD KLD Loss         & 86.26 & 88.83 & 93.70 & 60.43 & 70.78 & 95.40 & 82.57 \\
KD Ulb              & 87.91 & 89.11 & 94.49 & 70.28 & 73.33 & 95.87 & 85.17 \\
KD Ulb + FastText   & 87.47 & 88.06 & 94.40 & 69.17 & 72.24 & 95.67 & 84.50  \\
KD Ulb + BERT Embed & 88.60 & 89.22 & 94.42 & 68.88 & 73.62 & 95.64 & 85.06 \\ 
\midrule 
BERT-Tiny & 67.53 & 78.60 & 89.83 & 34.06 & 48.14 & 93.02 & 68.53 \\
KD        & 83.32 & 87.54 & 92.82 & 47.83 & 68.21 & 93.97 & 78.95 \\
% KD KLD Loss & 78.24 & 84.82 & 92.02 & 34.63 & 65.58 & 93.62 & 74.82\\
KD Ulb & 86.38 & 87.21 & 94.56 & 64.65 & 71.87 & 95.37 & 83.34 \\
KD Ulb + FastText & 86.33 & 87.60 & 94.16 & 64.02 & 70.59 & 95.12 & 82.97 \\
KD Ulb + BERT Embed & 86.49 & 87.80 & 94.44 & 65.51 & 71.39 & 95.54 & 83.53 \\
\midrule
BERT-Mini & 72.10 & 79.97 & 90.95 & 34.89 & 56.69 & 93.05 & 71.27 \\
KD        & 88.41 & 88.62 & 94.21 & 55.59 & 72.78 & 95.38 & 82.50 \\
% KD KLD Loss & 83.04 & 85.98 & 92.22 & 51.91 & 67.65 & 94.22 & 79.17 \\
KD Ulb & 89.69 & 89.01 & 95.15 & 70.48 & 75.24 & 95.87 & 85.91 \\
KD Ulb + FastText & 89.86 & 89.34 & 95.28 & 70.73 & 75.31 & 95.95 & 86.08 \\
KD Ulb + BERT Embed & 90.20 & 89.14 & 95.23 & 71.72 & 76.16 & 96.05 & 86.42 \\
\midrule
BERT-Small          & 75.89 & 80.03 & 91.32 & 32.88 & 56.62 & 93.28 & 71.67 \\
KD                  & 88.27 & 88.33 & 94.18 & 60.37 & 72.05 & 95.14 & 83.06 \\
% KD KLD Loss         & 84.16 & 86.92 & 93.02 & 55.06 & 69.09 & 94.70 & 80.49 \\
KD Ulb              & 89.65 & 88.82 & 95.21 & 71.11 & 75.20 & 95.87 & 85.98 \\
KD Ulb + FastText   & 90.00 & 88.70 & 95.37 & 70.86 & 75.96 & 96.02 & 86.15 \\
KD Ulb + BERT Embed & 90.65 & 88.97 & 95.32 & 72.58 & 76.52 & 96.10 & 86.69 \\ 
\bottomrule
\end{tabular}
\end{table*}

\subsubsection{Baseline}

Before performing KD from the teacher model to each of the student models, we first explore each ``student'' performance under normal training condition without KD. We also train the ``teacher'' model, that is, the BERT-base model, to see the best performance a dataset can achieve.

For text classification, BERT-Small gets the best performance among the student models with an F1 score of 80.78, followed by BERT-Mini (see Table~\ref{tbl-cls}). It reaffirms the power of Transformer-based models. Among the others, we find that CNN performs better than BiLSTM in the base settings.

Unlike the classification task, in sequence labeling, all BERT variant models perform worse than BiLSTM and CNN. Even the BERT-Small's performance is lower, which is interesting. We assume that it is because BERT models need a larger amount of training data but our sequence labeling datasets are relatively small. Between the non-Transformer models, BiLSTM is slightly better than CNN.

For both classification and sequence labeling, we see the upwards trend in BERT's variant. The bigger the model, the better its performance, similar to~\citet{turc2019}'s reported results.

\subsubsection{Knowledge Distillation}

After having the baseline, we continue with the KD experiment, using a fine-tuned BERT-base model as the teacher for each dataset. From this experiment, we find that all student models gain some improvement after training with KD compared to the baseline vanilla training. 

In the classification task, BiLSTM gets the most gain with 5.5 points of improvement, followed by BERT-Mini with 4.88. These two models are almost as good as BERT-Small with KD. If we consider the model's size and inference time (discussed more in Section~\ref{sec:cost-analysis}), these methods can be more appealing to be used.

In sequence labeling, BERT-Small and BERT-Mini get the most gains with more than 11 points of improvement. For non-BERT variant, CNN is slightly better than BiLSTM after KD with around 7 points of improvement. 

We also performed a small experiment to compare MSE vs KLD, where the former outperforms the latter by 1.66 points on average, further confirming \citet{KimOKCY21comparing} results.

% In sequence labeling task, we compare two different loss function: MSE and KLD as several previous works use KLD for their loss function when they work on sequence labeling data. Although both loss function improve the performance from the baseline, we found that it depends on the method itself in which loss function would gave better result.
% In BiLSTM, using KLD is always better than MSE, but in BERT-Small, MSE always outperform KLD. In CNN, several dataset performed better when using MSE while others using KLD is better, but the overall performance show that MSE is better. In BERT variants, using MSE is consistently better than KLD.
% Because of the difference in result, in the next step, for each model we use the best loss function based on their previous result.

\subsubsection{Unlabelled Data}
\label{sec-ulb-class}

KD naturally enables data augmentation because we can just use the result from the teacher's model when fed with unlabeled data, instead of using real labeled data. In our experiment, our unlabeled data comes from all datasets for each task, merged into one, without considering the original labels.
For classification, the total amount of unlabelled data is 74688 whereas for sequence labeling it is 30130.
Note that some data would certainly be out-of-domain. As such, we can directly observe the effect of out-of-domain data on the KD's performance.

On average, adding more data increases the overall model performance for both classification and sequence labeling tasks, despite the possible out-of-domain issues.
For classification, Bi-LSTM improves the most in comparison to the regular KD with 1.51 points of improvement.
For sequence labeling, Bert-Mini gains the largest improvement with 3.39 points, almost reaching the performance of BERT-Small with the same settings. 

An interesting case happened for CNN in the classification task as adding unlabelled data worsens the final performance. We assume that adding many unlabelled data increases noise and could create data imbalance, for which the CNN model is more sensitive to, thus resulting in a worse performance. 
We further explore this in Section~\ref{sec-discuss}.

\subsubsection{Embeddings}

We want to explore the effect of initializing the embedding layer of our student models with pre-trained embeddings. We experimented with two types of embedding: non-contextual embedding using FastText and contextual embedding using pre-trained BERT.

\begin{figure}[h]
  \caption{Number of step it is needed for the training to converge between using pretrained embedding (green) and not (blue).}
  \includegraphics[width=\linewidth]{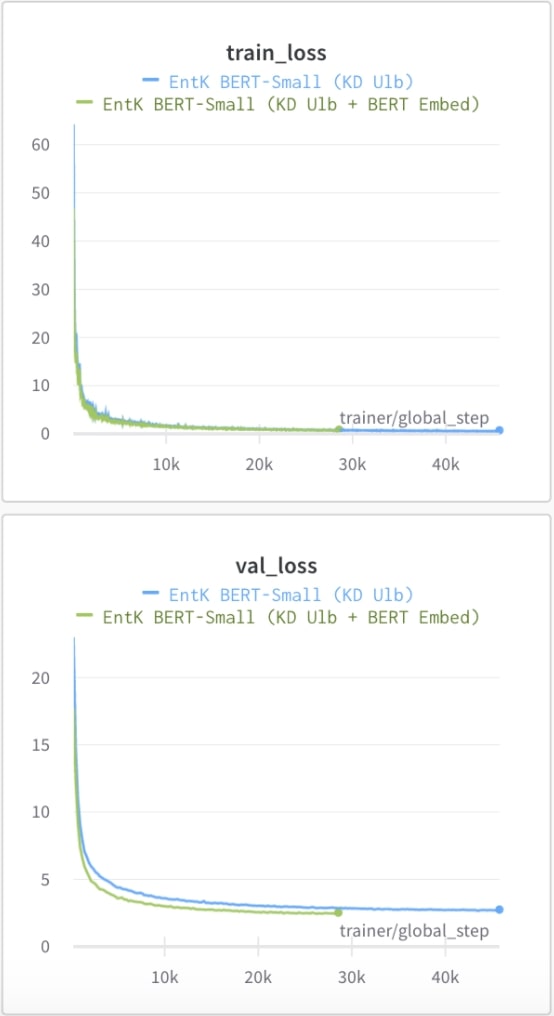}
  \label{fig:train-converge-embed}
\end{figure}

On average, our results show that using BERT embeddings gives more improvement compared to FastText. At its best, it was just one to two points below the performance of the original teacher model. Though the improvements seem to be relatively small when compared to the gains from data augmentation, using pretrained embeddings also helps the models converge faster (see Figure~\ref{fig:train-converge-embed}).

\subsection{Cost Analysis}
\label{sec:cost-analysis}
Though the F1 scores give a clear understanding of the model performance, as each student model has different sizes and complexity, there are some trade-offs. To give a deeper analysis, we compare each model size and its inference time to provide valuable insight towards model deployment. Table~\ref{tbl-cost} contains our results.

\begin{table}[width=.9\linewidth,cols=5,pos=h]
\caption{Model Size and Inference time comparison. Model size is in number of million (M) of parameters. Inference Time for both CPU and GPU are in milliseconds (ms).}
\label{tbl-cost}
\begin{tabular}{lrrr}
\toprule
\multicolumn{1}{c}{Model} & \multicolumn{1}{c}{Param} & \multicolumn{1}{c}{CPU} & \multicolumn{1}{c}{GPU} \\ \midrule
\multicolumn{4}{c}{Classification} \\ \midrule
BERT & 124.51 & 39.15 & 11.69 \\
Bi-LSTM & 3.35 & 1.74 & 1.09 \\
CNN & 4.02 & 2.36 & 2.96 \\
BERT-Tiny & 4.40 & 2.07 & 2.47 \\
BERT-Mini & 11.19 & 4.52 & 4.21 \\
BERT-Small & 28.81 & 8.93 & 4.28 \\ 
\midrule
\multicolumn{4}{c}{Sequence Labeling} \\ \midrule
BERT & 124.46 & 67.18 & 12.43 \\
Bi-LSTM & 4.38 & 3.19 & 1.61 \\
CNN & 4.01 & 2.35 & 3.12 \\
BERT-Tiny & 4,40 & 2.24 & 2.46 \\
BERT-Mini & 11.18 & 6.32 & 4.48\\
BERT-Small & 28.78 & 9.25 & 4.50 \\ 
\bottomrule
\end{tabular}
\end{table}

\subsubsection{Model Size}

On average, the older deep learning models like BiLSTM and CNN are 31x smaller with only 3-4 F1 score difference in its best KD setting compared to the teacher model.
If we want to lessen the loss of performance, BERT-small and BERT-Mini are viable options, with 1-2 and 2-3 points below the teacher model, respectively, but at the cost of being up to 7x bigger than CNN and BiLSTM. When we need to store multiple versions of a model from various cases, a smaller-sized model could significantly save up space.

\subsubsection{Inference time}

As expected from deep neural models, using GPUs can speed up the inference time. An exception occurred for convolutional models that used depthwise separable convolution operations (like our CNN models), which is not a hardware issue but a known software issue related to the CUDA library\footnote{\url{https://github.com/pytorch/pytorch/issues/18631}}.  

From Table~\ref{tbl-cost} we can see that for most of the student models, the difference of using CPU and GPU is not that big, except for BERT-Small as it is the largest model.
Bi-LSTM and CNN perform faster as it also has the least amount of parameter. BERT-Tiny also runs as fast as these two models but has the worst performance, and thus should never be considered. On the other hand, the better performing models like BERT-Small and BERT-Mini come with speed penalty, with each running 2-4x slower than BiLSTM/CNN, while only outperforming them in terms of F1 by 2-3 points on average. When scalability becomes an issue, it might be beneficial to opt for the smaller BiLSTM or CNN models instead.

\subsubsection{Using ONNX}

ONNX is a cross-platform inferencing and training accelerator compatible with popular ML/DNN frameworks, including PyTorch, TensorFlow/Keras, scikit-learn, and more.\footnote{\url{https://onnx.ai/}} It can be executed in different runtime environments and does not need to package every dependency. 

ONNX Runtime has been shown to speed up model inference time on a CPU machine without any loss of accuracy~\citep{wu_2020,alluin_2021}. As a CPU server is typically cheaper and more available than a GPU server, it is often beneficial to provide an optimized model version that runs well on the CPU. Converting the model to ONNX is also shown to remove PyTorch library dependency, which results in a smaller model image size.

From Table~\ref{tbl-onnx} we can see that using ONNX reduces the model size and in most cases, speeds up the inference time. On average, it reduces up to 60\% of the original model size.
For the student models, converting to ONNX also boost its inference time in CPU. For example, our BiLSTM ONNX model becomes 4x faster than the original performance, which is also 20x faster than the ONNX BERT-Small model. This further highlights the competitive advantages of the smaller BiLSTM and CNN student models over the BERT-Small model.

\begin{table}[width=.9\linewidth,cols=5,pos=h]
\caption{Comparison between using original model and ONNX version. Model size in according to its file size in MB and inference time using CPU in milliseconds (ms).}
\label{tbl-onnx}
\begin{tabular}{lrrrr}
\toprule
\multicolumn{1}{c}{\multirow{2}{*}{Model}} & \multicolumn{2}{c}{Original} & \multicolumn{2}{c}{ONNX} \\
\multicolumn{1}{c}{} & \multicolumn{1}{c}{Size} & \multicolumn{1}{c}{CPU} & \multicolumn{1}{c}{Size} & \multicolumn{1}{c}{CPU} \\ \midrule
% \multicolumn{5}{c}{Classification} \\
% \midrule
BERT &  1425.17 & 39.15 & 475.06 & 46.72 \\
Bi-LSTM & 38.40 & 1.74 & 12.80 & 0.36 \\
CNN & 46.06 & 2.36 & 15.35 & 0.34\\
BERT-Tiny & 50.38 & 2.07 & 16.79 & 0.30 \\
BERT-Mini & 128.20 & 4.52 & 42.73 & 1.78 \\
BERT-Small & 329.81 & 8.93 & 109.94 & 7.25 \\ 
% \midrule
% \multicolumn{5}{c}{Sequence Labeling} \\
% \midrule
% BERT & \\
% Bi-LSTM & \\
% CNN & \\
% BERT-Tiny & \\
% BERT-Mini & \\
% BERT-Small & \\
\bottomrule
\end{tabular}
\end{table}

\section{Discussion}
\label{sec-discuss}

In this section, we discuss more about some important caveats on our KD results, especially in some places where our pipeline via unlabeled data augmentation or embeddings worsens the results.

\subsection{Hyperparameter Tuning}

For some datasets on CNN student model, we find that some of our pipeline (unlabeled data augmentation or embeddings) does not increase the performance of our KD. We then seek to improve the results by performing hyperparameter optimization. Specifically, we focus on optimizing the learning rate via random search between [5e-5, 1e-2]. From this small experiment, we obtain the best performance for Health dataset (F1 Score=74.68) when using learning rate of 8e-4 and for Telco dataset (F1 score=87.74) when using learning rate of 2e-3. Both learning rates are different from the ones being used in the default experiment setup. It shows the importance of performing hyperparameter tuning for several datasets. 

\begin{figure}[ht]
    \caption{F1 score with different learning rate for health and telco dataset.}
    \label{fig-lr-tune}
    \centering
    \begin{tikzpicture}
        \begin{axis}[%
        scatter/classes={%
            health={mark=x, draw=red},
            telco={mark=o, draw=blue}
        },
        xlabel=learning rate,
        width=220pt, height=195pt
        ]
        \addplot[scatter,only marks, mark size=2.5pt,
            scatter src=explicit symbolic]%
        table[meta=label] {
        x y label
        0.00005 68.34 health
        0.0001 70.04 health
        0.0005 74.26 health
        0.0008 74.58 health
        0.0008 78.75 telco
        0.001 74.32 health
        0.001 77.77 telco
        0.002 87.74 telco
        0.005 72.94 health
        0.005 84.05 telco
        0.008 75.16 telco
        0.01 66.60 telco
        };
        \legend{health, telco}
        \end{axis}
    \end{tikzpicture}
\end{figure}
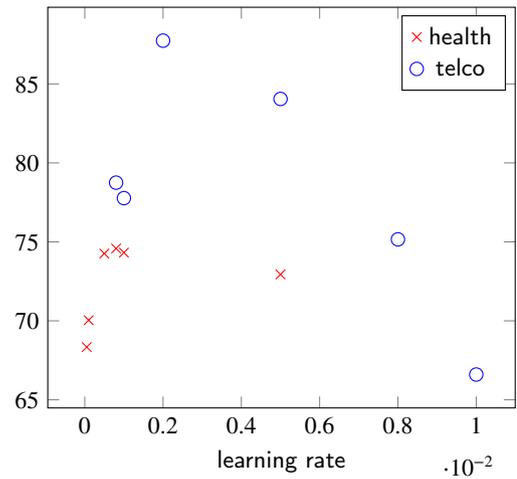

\subsection{Out-of-Domain Data}
\label{sec-ood}

As we state in Section~\ref{sec-ulb}, our unlabelled data augmentation comes from all datasets combined, which gives rise to the possibility of domain mismatch. We suspect this may worsen the performance for several KD with unlabelled data. For example, in Telco dataset, the performance for CNN ulb is worse than the baseline and BiLSTM KD-ulb only received a small gain in comparison to the vanilla KD. To test this hypothesis, we create two alternative scenarios to add new unlabeled data into the Telco dataset.

The first scenario is to perform data augmentation via back translation (BT), that is, by translating the Indonesian text data into English and then back to Indonesian using Google Translate, and then by randomly substituting some words by masking them and using some Indonesian BERT-based PLMs to predict the masked words. For the second scenario, we collect a new unlabeled dataset from Twitter which mentions the handle of some Indonesian telecommunication companies, thus ensuring the unlabeled data is within the same domain as the Telco dataset. We obtain 20000 new data for each scenario and combine each with the original Telco training dataset, resulting in 31520 text data for each scenario.

\begin{table}[width=.9\linewidth,cols=5,pos=h]
\caption{Telco dataset with in-domain unlabelled data.}
\label{tbl-ulb-indomain}
\begin{tabular}{lrrr}
\toprule
\multicolumn{1}{c}{data telco} & \multicolumn{1}{c}{BiLSTM} & \multicolumn{1}{c}{CNN} & \multicolumn{1}{c}{N} \\ \midrule
% KD & 91.41 & 90.95 \\
ulb & 92.71 & 84.05 & 74688 \\ 
% \midrule
% ulb-data aug & & 92.19 & 51853 \\ 
% ulb-in domain & & 86.10 & 32864 \\ 
\midrule
BT+subs & 93.07 & 91.96 & 31520 \\ 
Twitter in-domain & 91.23 & 84.45 & 31520 \\
\bottomrule
\end{tabular}
\end{table}

From Table~\ref{tbl-ulb-indomain} we can see that the back-translation plus substitution scenario gives a better performance, especially for the CNN model which improves it by 7.91\%. The second scenario which uses new Twitter data works as well as the original ulb strategy, but does not really improve it. This indicates that sometimes the domain of the unlabeled data matters, and that CNN student models might be more sensitive towards it. Using strategies that keep the augmented data as close as possible to the original data also seem to work well, as proven by the back-translation and substitution strategy.

\subsection{Data Imbalance}
\label{sec-imbalance}

Having imbalanced data for each label often makes training more difficult. Adding more unlabelled data might exacerbate this issue if the numbers of data generated per label are not controlled. We hypothesize that this might be the source of some worse results after applying the unlabeled data augmentation. 

To validate this, we use again the Telco dataset. The original dataset contains a balanced label distribution with 144 labels having 80 training data each, resulting in a total of 11520 data. The testing and validation are 10 data for each label. With the addition of unlabelled data, this balance was disrupted. To ensure the balance is kept, we do the following.

First, we analyze the statistics of the labels of the unlabelled data. Using the teacher model, we run predictions to those data to know the resulting label distributions. We find that the data is highly imbalanced. There are labels with very few (0-10) additional data, while there are several others with very large (> 1000) additional data (with a maximum of 11050). The median additional data per label is 154. Next, to make it more balanced, for each label we randomly sample at most 154 additional unlabelled data, following the median. In this scenario, the total number of unlabeled data that we use is 26199.

With 154 additional data per label, the dataset is now balanced but the total number is much lower than the original augmentation which results in an average of 518.67 data per label in the Telco dataset. Therefore, we create a second scenario where we randomly sample 518 data per label. For labels that have less than 518 data, we perform oversampling so that they have 518 but with some duplicates.

\begin{table}[width=\linewidth,cols=5,pos=h]
\caption{N=total unlabeled data, std=standard deviation of the numbers of augmented data per label. Lower std means more balanced data. The oversampling (-o) forces all labels to have exactly the same numbers of data each, hence the 0 std.}
\label{tbl-ulb-balance}
\begin{tabular}{lrrr}
\toprule
% lr    & kd-ulb & kd-ulb-balance \\ \midrule
% 0.002 & 87.74  & 82.06     \\
% 0.005 & 84.05  & 83.27     \\ 
Telco data augmentation & CNN (F1) & N & std \\ \midrule
ulb & 84.05 & 74688  & 1234.33 \\
\midrule
ulb-median & 83.27 & 26199 & 58.99 \\
ulb-mean-o & 80.68 & 74592 & 0.00 \\ \midrule
ulb-mean-o min.max & 88.34 & 70139 & 55.14 \\ 
\bottomrule
\end{tabular}
\end{table}

% telco
% mean	6.203038
% std	2.646163
% min	1.000000
% 25%	4.000000
% 50%	6.000000
% 75%	8.000000
% max	24.000000

% ulb-balance
% count	74592.000000
% mean	9.280955
% std	9.843315
% min	1.000000
% 25%	5.000000
% 50%	7.000000
% 75%	9.000000
% max	128.000000

% ulb-balance min.max
% count	70139.000000
% mean	7.139594
% std	3.780981
% min	1.000000
% 25%	5.000000
% 50%	6.000000
% 75%	9.000000
% max	24.000000

Based on Table~\ref{tbl-ulb-balance}, we see that balancing the data does not improve influence KD training, but the opposite. Additionally, we find that filtering the augmented data based on length, so that only those between the minimum and the maximum length of the original Telco data are considered, improves the results. We explore this in the next subsection below.

\subsection{Sentence length}

BiLSTM is inherently designed to model long sequences, unlike CNN. Therefore, we suspect that the length of the text itself may play some part in worsening the CNN KD results after applying unlabeled data augmentation. Several datasets like Telco and Smltk contain short sentences and may be contaminated with longer sentences when augmented with other datasets.

To validate this, we use the Smltk dataset. We analyze the original sentence length of the Smltk dataset and the unlabelled dataset (without Smltk) using BERT Tokenizer. The original Smltk data has on average 6.4 tokens per text with std of 2.88. The minimum is 1 and the maximum is 21. The first quartile ($Q_1$) is 4 while the third quartile ($Q_3$) is 8. 
On the other hand, the unlabelled data (without smltk) has an average of 14.43, std if 15.88, min of 1, max of 130. We can immediately see the discrepancies between them in terms of length. Therefore, we aim to filter the unlabelled data so that it better matches the length of the Smltk data. We do this by taking only the texts with the length within a certain range. We test two ranges: (1) following the minimum and maximum of the Smltk data, and (2) following the $Q_1$ and $Q_3$ of the Smltk data. The filtered unlabeled dataset from each of the above scenarios is then combined with the original Smltk data. Table~\ref{tab:smltk-stat} shows the statistics of the resulting datasets, with
both the two new datasets have a closer distribution in terms of sentence length to the original smalltalk dataset. 

\begin{table}[width=.9\linewidth,cols=5,pos=h]
\label{tab:smltk-stat}
\caption{Smltk dataset statistic based on tokens length}
\label{tbl-ulb-senlen-dist}
\begin{tabular}{lrrrrrr}
\toprule
\multicolumn{1}{c}{data} & \multicolumn{1}{c}{mean} & \multicolumn{1}{c}{std} & \multicolumn{1}{c}{min} & \multicolumn{1}{c}{$Q_1$} & \multicolumn{1}{c}{$Q_3$} & \multicolumn{1}{c}{max} \\ \midrule
smltk& 6.40 & 2.88& 1 & 4 & 8 & 21 \\
ulb (- smltk) & 15.81& 16.78 & 1 & 5 & 18 & 130 \\
\midrule
ulb-min.max & 8.00 & 4.47& 1 & 5 & 10 & 21 \\
ulb-$Q_1$.$Q_3$ & 6.07 & 1.99& 1 & 5 & 7 & 21 \\ 
\bottomrule
\end{tabular}
\end{table}

\begin{table}[width=.9\linewidth,cols=5,pos=h]
\caption{Smltk data F1 with unlabelled dataset comparison}
\label{tbl-ulb-senlen}
\begin{tabular}{lrr}
\toprule
\multicolumn{1}{c}{data smltk} & \multicolumn{1}{c}{BiLSTM} & \multicolumn{1}{c}{CNN} \\ \midrule
% KD & 92.96 & 92.43 \\
ulb & 92.74 & 84.50 \\ 
\midrule
ulb-min.max & 91.53 & 91.66 \\ 
ulb-$Q_1$.$Q_3$ & 92.87 & 92.30 \\ 
\bottomrule
\end{tabular}
\end{table}

Using the new unlabeled data, we retrain again the student models. The result is in Table~\ref{tbl-ulb-senlen}, from which we can see that sequence length matters a lot, especially for the CNN model.

Knowing these, the results of the domain-dependence test using back-translation and substitution (Section~\ref{sec-ood}) might be because the augmented data have similar token length to the original dataset, while the Twitter scraping approach might result in long texts. Other than that, we have seen that balancing the data (Section~\ref{sec-imbalance}) with no attention to the token's length fails to improve the result. We leave exploration into the interplay between these three factors in data augmentation as future work.

\section{Conclusion and Future Work}
We explored task specific distillation using Transformer based teacher model and several student models. Throughout our experiment, we have shown that some student models, BiLSTM and CNN, yield the best results when taking into account both F1-score and computational cost. When considering model compression focusing on scalability, using either of the two models for KD over small BERT models is recommended, as the performance gain on the BERT models do not scale-up with the computational burden. Practically, we recommend designing a cascading system, i.e. start your NLP service with the BERT teacher model, then downgrade to BERT-small, BERT-Mini, and BiLSTM/CNN when the service demand is rising. A separate model can be trained to track the demand fluctuations throughout days, weeks, months, or years, so that the best model at a given timeframe is chosen.

We additionally recommend some KD quick-wins, i.e., effective KD improvements with little-to-no drawbacks in terms of computational cost:
\begin{enumerate}
    \item use MSE over KLD,
    \item use pretrained word embeddings for student initialization, with higher priority on BERT embeddings compared to Fasttext, and
    \item perform unlabeled data augmentation during KD.
\end{enumerate}

Our student models were limited to BiLSTM with attention, CNN, and some scaled-down BERT models. We recommend further exploration into specifically designed student models such as TinyBERT~\citep{jiao-etal-2020-tinybert}, or efficient variants of Transformer such as Nystromformer~\citep{xiong2021nystromformer} in the future because they might provide better performance-computational trade-offs. For the word embeddings, we recommend exploration into the deeper layer of transformer-based models. For example, going into the first transformer block of a pretrained BERT model as the ``embedding layer''. This certainly will make the model bigger and slower, but the performance gain might be worth it.

Though data augmentation works almost across the board, we recommend preparing data with similar length compared to the original labeled data, as some student models (especially CNN) are sensitive to input length variations. We also recommend augmenting data within the intended domain, while we also recommend against balancing the data for KD purposes. Though, results on this were inconclusive and might be affected by the text length itself, so we recommend further exploration in this direction.

Our hyperparameter tuning experiment shows promising results on the CNN model. We recommend using more computational power for this purpose instead of for deploying BERT-base models, as the final performance of the tuned CNN (or LSTM) student models can potentially reach the BERT models. Finally, we highly recommend transforming the student models into ONNX runtime as it further boosts the running time and model size.

\section*{Acknowledgement}

This research is fully and exclusively funded by PT. Yesboss Group Indonesia (Kata.ai), where the authors work.

%We performed on two tasks, text classification and sequence labeling, with Indonesian dataset. We also explore several ways to improve KD's result and investigate why it didn't performed well on several dataset and algorithms.

\printcredits

\bibliographystyle{cas-model2-names}

\bibliography{cas-refs}


\end{document}